\documentclass[letterpaper, 10 pt, conference]{ieeeconf}  

\IEEEoverridecommandlockouts                              

\usepackage{graphics} 
\usepackage{graphicx}
\usepackage{amsmath} 
\usepackage{amssymb}  
\usepackage{color}
\usepackage{cite}
\usepackage{hyperref}

\usepackage{multirow}
\usepackage{algorithm,algcompatible}
\usepackage{bm}
\usepackage[tight,footnotesize]{subfigure}
\usepackage{multicol}

\algnewcommand\INPUT{\item[\textbf{Input:}]}%
\algnewcommand\OUTPUT{\item[\textbf{Output:}]}%

\newcommand{\etal}{\textit{et al}.}
\newcommand{\uvec}[1]{\boldsymbol{\hat{\textbf{#1}}}}

\newcounter{algsubstate}

\title{\LARGE \bf
Visual Odometry Revisited: What Should Be Learnt?
}

\author{
Huangying Zhan, 
Chamara Saroj Weerasekera,
Jia-Wang Bian,
Ian Reid
\thanks{All authors are with the School of Computer Science, at the University
of Adelaide, and Australian Centre for Robotic Vision}%
}

\begin{document}
\maketitle
\thispagestyle{empty}
\pagestyle{empty}
\begin{abstract}
In this work we present a monocular visual odometry (VO) algorithm which leverages geometry-based methods and deep learning. 
Most existing VO/SLAM systems with superior performance are based on geometry and have to be carefully designed for different application scenarios. 
Moreover, most monocular systems suffer from scale-drift issue.
Some recent deep learning works learn VO in an end-to-end manner but the performance of these deep systems is still not comparable to geometry-based methods.
In this work, we revisit the basics of VO and explore the right way for integrating deep learning with epipolar geometry and Perspective-n-Point (PnP) method.
Specifically, we train two convolutional neural networks (CNNs) for estimating single-view depths and two-view optical flows as intermediate outputs. 
With the deep predictions, we design a simple but robust frame-to-frame VO algorithm (DF-VO) which outperforms pure deep learning-based and geometry-based methods.
More importantly, our system does not suffer from the scale-drift issue being aided by a scale consistent single-view depth CNN.
Extensive experiments on KITTI dataset shows the robustness of our system and a detailed ablation study shows the effect of different factors in our system. 
Code is available at here:  \href{https://github.com/Huangying-Zhan/DF-VO}{DF-VO}.
\end{abstract}

\section{Introduction} \label{Sec:intro}
The ability for an autonomous robot to know its whereabouts and its surroundings is of utmost importance for tasks such as object manipulation and navigation. Vision-based localisation and mapping is often the preferred choice due to factors such as low cost and power requirements, and it can provide useful complementary information to other sensors such as IMU, GPS, laser scanners, etc.
Two broad types of visual localisation methods are: Visual Odometry -- the main focus area of this work -- and Simulataneous Localisation and Mapping (SLAM). VO is useful when the (6DoF) motion of the robot relative to its previous state is of main interest, while visual SLAM is more suited when an accurate robot trajectory and map of the environment are required, and the latter also involves closing loops and re-localizing the robot when tracking is lost and it is re-visiting the same environment again. 
Moreover, VO is an integral part of SLAM for inter-frame tracking, and more accurate odometry reduces overall drift in localisation, and minimises the need for loop closure and global refinement of the camera trajectory.

Pure multi-view geometry-based VO is reliable and accurate under favourable conditions, such as when there is sufficient illumination and texture to establish correspondence \cite{lowe2004sift, rublee2011orb, Bian2019gms}, sufficient overlap between consecutive frames, and when majority of the visible scene is static (few moving objects).
\begin{figure}[t!]
        \centering
		\includegraphics[width=1\columnwidth]{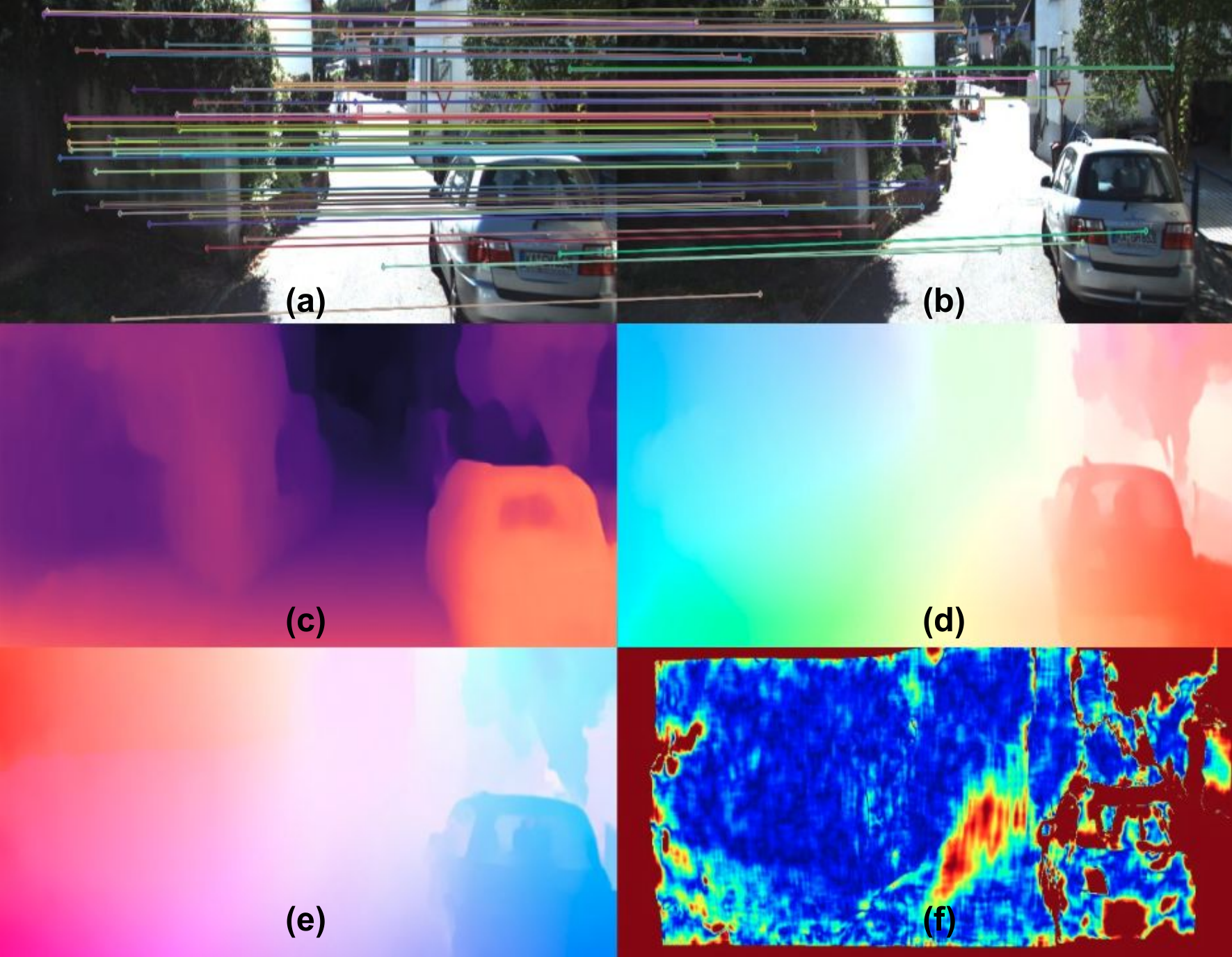}
		\vspace{-10pt}
		\caption{
		Inputs and intermediate CNN outputs of the system. 
		(a, b) Current and previous monocular input images with examples of auto-selected 2D-2D matches; 
		(c) Single-view depth prediction;
		(d, e) Forward and backward optical flow prediction;
		(f) Forward-backward flow consistency using (d,e).
		}
		\label{fig:vis_eg}
		\vspace{-8pt}
\end{figure}
Having an accurate estimation of translation scale per-frame is crucial for any monocular VO/SLAM system. 
However, pure geometry-based VO/SLAM suffer from the scale drift issue. 
Resolving scale drift usually relies on keeping a scale consistent map for map-to-frame tracking, performing an expensive global bundle adjustment for scale optimization or additional prior assumptions like constant camera height from the known ground plane.
Moreover, most monocular systems suffer from a single depth-translation scale ambiguity issue, which means the predictions (structure and motion) are up-to-scale.

Recently deep learning based methods have made possible end-to-end learning of camera motion from videos \cite{zhou2017sfmlearner,zhan2018depthVO,yin2018geonet, ranjan2019cc,bian2019depth}, or prediction of pose with respect to a scene from a single view \cite{kendall2015posenet, brachmann2017dsac, brachmann2018dsac2}.
Also, learning from data allows the deep networks to give predictions associated with real world scale, which solves the scale ambiguity issue.
While these learnt systems enable camera tracking/localisation in challenging conditions, these systems fail to provide the reliablility and accuracy of pure geometry based methods in those conditions that favour geometry-based algorithms. 

In this work we revisit the basics of geometry-based VO and explore the right way of incorporating deep learning into it.
A simple and robust frame-to-frame VO algorithm, named DF-VO, incorporating deep predictions (Fig. \ref{fig:vis_eg}) is proposed.
We extensively compare our system against both deep learning methods and geometry methods.
Moreover, we conduct a detailed ablation study for evaluating the effect of different factors in our system.

\section{Related Work} \label{Sec:rel_work}
\textbf{Geometry based VO: }
Camera tracking is a fundamental and well studied problem in computer vision, with different pose estimation methods based on multiple-view geometry been established \cite{HartleyMultiView}\cite{scaramuzza2011visual}. 
Early work in VO dates back to the 1980s \cite{ullman1979motion}\cite{scaramuzza2011visual}, with a successful application of it in the Mars exploration rover in 2004 \cite{matthies2007computer}, albeit with a stereo camera. Two dominant methods for geometry based VO/SLAM are feature-based \cite{ORBSLAM2, klein2007ptam, Geiger2011viso2} and direct methods \cite{DSO,newcombe2011dtam}. The former involves explicit correspondence estimation, and the latter takes the form of an energy minimisation problem based on the image color/feature warp error, parameterized by pose and map parameters. There are also hybrid approaches which make use of the good properties of both \cite{forster2014svo,forster2016svo,engel2014lsd}.
One of the most successful and accurate full SLAM systems using a sparse (ORB) feature-based approach is ORB-SLAM2 \cite{ORBSLAM2}, along with DSO \cite{DSO}, a direct keyframe-based sparse SLAM method. VISO2 \cite{Geiger2011viso2} on the other hand is a feature-based VO system which only tracks against a local map created by the previous two frames. All of these methods suffer from the previously mentioned issues (including scale-drift) common to monocular geometry-based systems, which we show our proposed method addresses.

\textbf{Deep learning for VO: }
For supervised learning, Agrawal \etal \cite{agrawal2015seebymoving} propose to learn good visual features from a ego-motion estimation task, in which the model is capable of relative camera pose estimation. 
Wang \etal \cite{wang2017deepvo} propose a recurrent network for learning VO from videos.
Ummenhofer \etal \cite{ummenhofer2016demon} and Zhou \etal \cite{deeptam} propose to learn monocular depth estimation and VO together in an end-to-end fashion by formulating structure from motion as a supervised learning problem.
Dharmasiri \etal \cite{dharmasiri2018eng} train a depth network and extend the depth system for predicting optical flows and camera motion.
Recent works suggest that both tasks can be jointly learnt in a self-supervised manner using a photometric warp loss to replace a supervised loss based on ground truth.
SfM-Learner \cite{zhou2017sfmlearner} is the first self-supervised method for jointly learning camera motion and depth estimation. 
SC-SfM-Learner \cite{bian2019depth} is a very recent work which solves the scale inconsistent issue in SfM-Learner by enforcing depth consistency. 
Some prior work like, UnDeepVO \cite{li2017undeepvo} and Depth-VO-Feat \cite{zhan2018depthVO}, solve the both scale ambiguity and inconsistency issue by using stereo sequences in training, which address the issue of metric scale.
\cite{yin2018geonet, ranjan2019cc} also incorporate optical flow in their joint training framework.

The issue with the above learning-based methods is that they don't explicitly account for the multi-view geometry constraints that are introduced due to camera motion \emph{during inference}. 
In order to address this, recent works have been proposed to combine the best of learning and geometry to varying extent and degree of success. CNN-SLAM \cite{tateno2017cnnslam} fuse single-view CNN depths in a direct SLAM system, and CNN-SVO \cite{loo2018cnnsvo} initialize the depth at a feature location with CNN provided depth for reducing the uncertainty in the initial map. 
Yang \etal \cite{yang2018dvso} feed depth predictions into DSO \cite{DSO} as virtual stereo measurements.
Li \etal \cite{li2019posevo} refine their pose predictions via pose-graph optimisation.
In contrast to the above methods, we effectively utilize CNNs for \emph{both} single-view depth prediction and correspondence estimation, on top of standard multi-view geometry to create a simple yet effective VO system.

\begin{algorithm} [t]
    \caption{DF-VO: Depth and Flow for Visual Odometry}
  \begin{algorithmic}[1]
    \REQUIRE Depth-CNN: $M_d$; Flow-CNN: $M_f$
    \INPUT Image sequence: $[I_1$, $I_2$, ..., $I_k]$
    \OUTPUT Camera poses: $[\bm{T_1}, \bm{T_2}, ..., \bm{T_k}]$
    \STATE \textbf{Initialization} $\bm{T_1} = \bm{I}$ ; $i=2$
    \WHILE{$i \leq k$}
        \STATE Get CNN predictions: $\bm{D}_i, \bm{F}^{i}_{i-1}, \text{and } \bm{F}^{i-1}_{i}$ 
        \STATE Compute forward-backward flow inconsistency.
        \STATE Form \textit{N}-matches $(\bm{P}_i, \bm{P}_{i-1})$ from flows with the least flow inconsistency.
        \IF{$\text{mean}(|\bm{F'})| > \delta_f$}
            \STATE Solve $\bm{E}$ from $(\bm{P}_i, \bm{P}_{i-1})$ and recover $[\bm{R}, \uvec{t}]$ 
            \STATE Triangulate $(\bm{P}_i, \bm{P}_{i-1})$ to get $\bm{D}'_{i}$
            \STATE Estimate scaling factor, $s$, by comparing $(\bm{D}_i, \bm{D}'_i)$
            \STATE $\bm{T}^{i-1}_{i} = [\bm{R}, s\uvec{t}]$
        \ELSE
            \STATE Form 3D-2D correspondences from $(\bm{D}_i, \bm{F}')$
            \STATE Estimate $[\bm{R}, \bm{t}]$ using PnP
            \STATE $\bm{T}^{i-1}_{i} = [\bm{R}, \bm{t}]$
        \ENDIF
        \STATE $\bm{T}_i \leftarrow \bm{T}_{i-1} \bm{T}^{i-1}_{i}$ 
    \ENDWHILE
  \end{algorithmic} \label{Alg:dfvo}
\end{algorithm}
%

\section{DF-VO: Depth and Flow for Visual Odometry}
\label{sec:method}
In this section, we revisit traditional pose estimation methods including epipolar geometry-based and PnP-based methods. 
We then follow up on the integration of these methods with our learned CNN outputs (Alg.\ref{Alg:dfvo}).

\subsection{Traditional Visual Odometry}\label{Sec:trad_VO}

\paragraph{Epipolar Geometry}
Given an image pair, $(I_1, I_2)$, the basic method for estimating the relative camera pose is solving fundamental or essential matrix, $\bm{E}$. 
When 2D-2D pixel correspondences $(\bm{p_1}, \bm{p_2})$ between the image pair are formed,
epipolar constraint is employed for solving the essential matrix. 
Thus, relative pose, $[\bm{R}, \bm{t}]$, can be recovered \cite{nister2003efficient,zhang1998determining,hartley1995defence,bian2019bench}.
\begin{align}
    \bm{p}_{2}^{T} \bm{K}^{-\bm{T}} \bm{E} \bm{K}^{-1} \bm{p}_{1} = 0, \text{where }  \bm{E} = [\bm{t}]_{\times}\bm{R} 
\end{align} 
where $\bm{K}$ is the camera intrinsics.
Typically, the 2D-2D pixel correspondences are formed either by extracting and matching salient feature points in the images, or by computing optical flow.
However, solving essential matrix for camera pose has some well-known issues.
\begin{itemize}
    \item Scale ambiguity: translation recovered from essential matrix is up-to-scale.
    \item Pure rotation issue: recovering $\bm{R}$ becomes unsolvable if the camera motion is pure rotation.
    \item Unstable solution: the solution is unstable if the camera translation is small.
\end{itemize}

\paragraph{PnP}
Perspective-n-Point (PnP) is a classic method for solving camera pose given 3D-2D correspondences.
Suppose the observed 3D points of view 1 and the observed projection in view 2 $(\bm{X_1}, \bm{p_2})$ are given, PnP can be deployed for solving the camera pose by minimizing the reprojection error,
\begin{align}
  e = \sum_{i} ||\bm{K} (\bm{R}\bm{X}_{(1,i)}+\bm{t}) - \bm{p}_{(2,i)}||_{2}
\end{align}
In order to establish the 3D-2D correspondences, we need to (1) estimate the 3D scene structure, (2) match 3D points to 2D pixels by matching features.

\subsection{Deep Predictions for Correspondence}
Suppose we have two deep models, $M_d$ and $M_f$, for single-view depth estimation and two-view optical flow estimation respectively. 
Given an image pair, $(I_{i}, I_{i-1})$, $M_f$ is employed to estimate the dense optical flow, which gives 2D-2D correspondences.
However, the CNN optical flows are not accurate for all the pixels. 
The accuracy of the 2D-2D correspondences is significant for estimating accurate relative pose.
In order to filter out the optical flow outliers,  
we estimate both forward and backward optical flows, $\bm{F}^{i}_{i-1}$ and  $\bm{F}^{i-1}_{i}$, and use the forward-backward flow inconsistency 
$|-\bm{F}^{i}_{i-1} - \bm{F}^{i-1}_{i}|$
as a measure to choose good 2D-2D correspondences.
We choose optical flows with the least flow inconsistency $\bm{F'}$ to form the best-\textit{N} 2D-2D matches, $(\bm{P}_i, \bm{P}_{i-1})$, for estimating the relative camera pose.
Comparing to traditional feature-based method, which only use salient feature points for matching and tracking, any pixel in the dense optical flow can be a candidate for tracking.
Moreover, traditional features usually gather visual information from local regions while CNN gathers more visual information (large receptive field) and higher level contextual information, which gives more accurate and robust estimations.
We can further estimate 3D structure w.r.t view-\textit{i} by estimating single-view depths from $M_d$. 
Knowing the 2D-2D matches and the 3D positions of view-\textit{i}, 3D-2D correspondences can be built.

\subsection{Deep Predictions for Visual Odometry}
Given a depth CNN prediction ($\bm{D}_i$) and $(\bm{P}_i, \bm{P}_{i-1})$, 2D-2D and 3D-2D correspondences can be established.
We can solve the relative camera pose either by solving PnP (3D-2D) or essential matrix (2D-2D).
Unfortunately, the current state-of-the-art (SOTA) single-view depth estimation methods are still insufficient for recovering accurate 3D structure for accurate camera pose estimation, 
which is shown in Table \ref{table:ablation}.

On the other hand, optical flow estimation is a more generic task and the current SOTA deep learning methods are accurate and with good generalization ability. 
Therefore, we can use the 2D-2D matches for solving essential matrix and recovering camera motion instead.
Nevertheless, as we mentioned in Sec.\ref{Sec:trad_VO}, solving essential matrix from two-view carries some well-known issues.
We show that we can avoid/resolve the issues by leveraging deep predictions.

\paragraph{Scale ambiguity} 
Translation recovered from essential matrix is up-to-scale.
However, we can use the CNN depths $\bm{D}_i$ as a reference for scale recovery.
First, $[\bm{R}, \uvec{t}]$ are recovered from solving essential matrix. 
Then, triangulation is performed for $(\bm{P}_i, \bm{P}_{i-1})$ to recover up-to-scale depths $\bm{D}'_i$. 
A scaling factor can be estimated by comparing $(\bm{D}_i, \bm{D}'_i)$. 

\paragraph{Unsolvable/Unstable solution}
Pure rotation causes unsolvable essential matrix while small camera motion is an ill-conditioned issue. In order to avoid these two issues, we apply two condition checks.
(1) \textit{Average flow magnitude}: we solve essential matrix only when the average flow magnitude is larger than a certain threshold,　$\delta_f = 5$, which avoids small camera motion which comes with small optical flows. 
(2) \textit{Check for cheirality condition}: There are 4 possible solutions for $[\bm{R}, \uvec{t}]$ by decomposing $\bm{E}$. 
In order to find the correct unique solution, cheirality condition, i.e. the triangulated 3D points must be in front of both cameras, is checked to remove the other solutions. 
We further use the number of points satisfying cheirality condition as a reference to determine if the solution is stable.

If the above conditions are not satisfied, the 3D-2D correspondences are used to estimate the camera motion by solving PnP instead.
To increase the robustness of the proposed pipeline, we wrap the following steps in RANSAC loops, including essential matrix estimation, scaling factor estimation, and PnP.

\section{CNN Training framework}3 
In this section, we present the deep learning networks and the training frameworks for learning depths and optical flows.

\subsection{Single-View Depth Network}
\begin{figure}[t!]
    \begin{multirow}{3}{*}
        \centering
        \begin{multicols}{2}
            \centering
            \includegraphics[width=1\columnwidth]{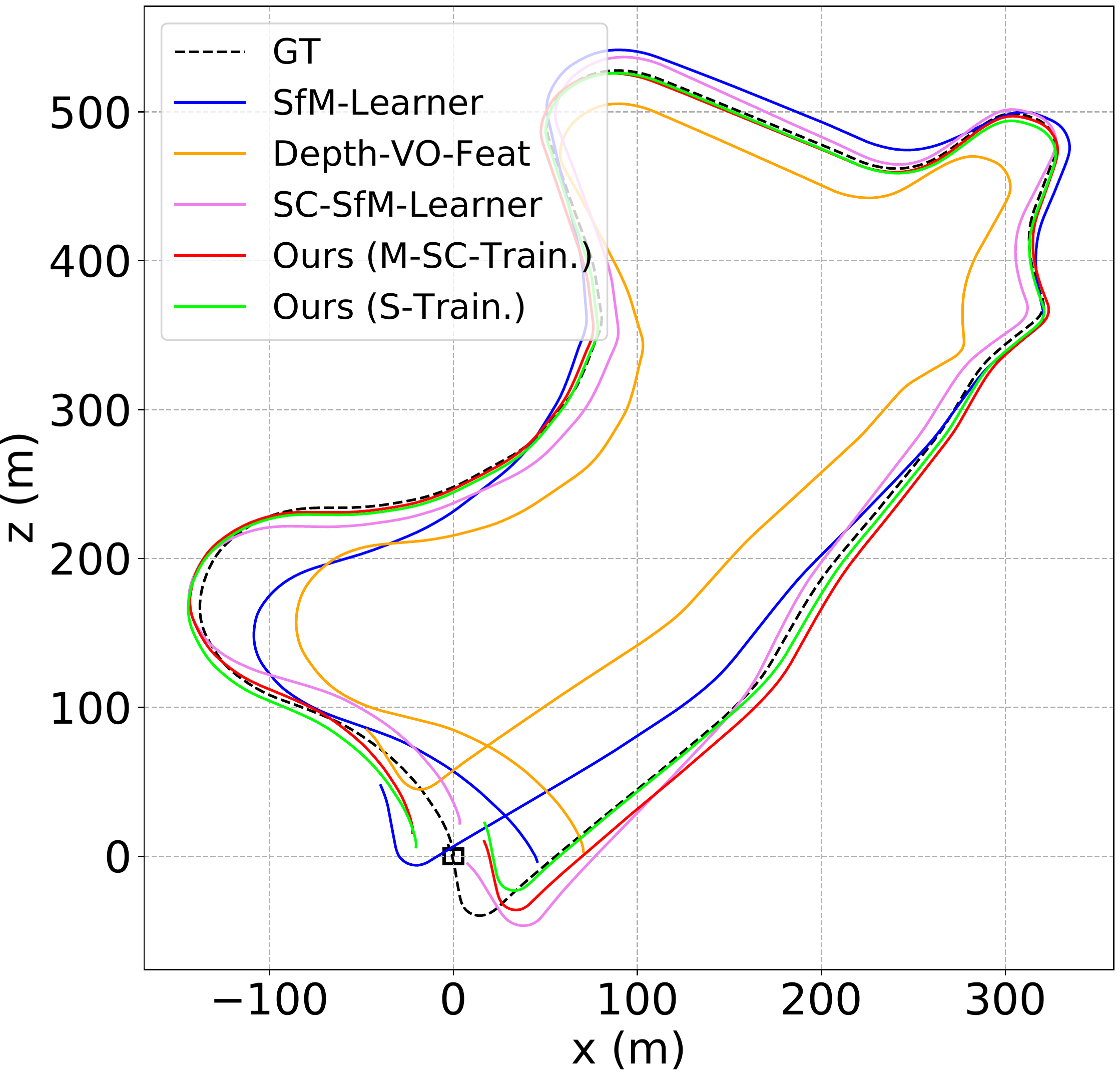}\\
            \includegraphics[width=1\columnwidth]{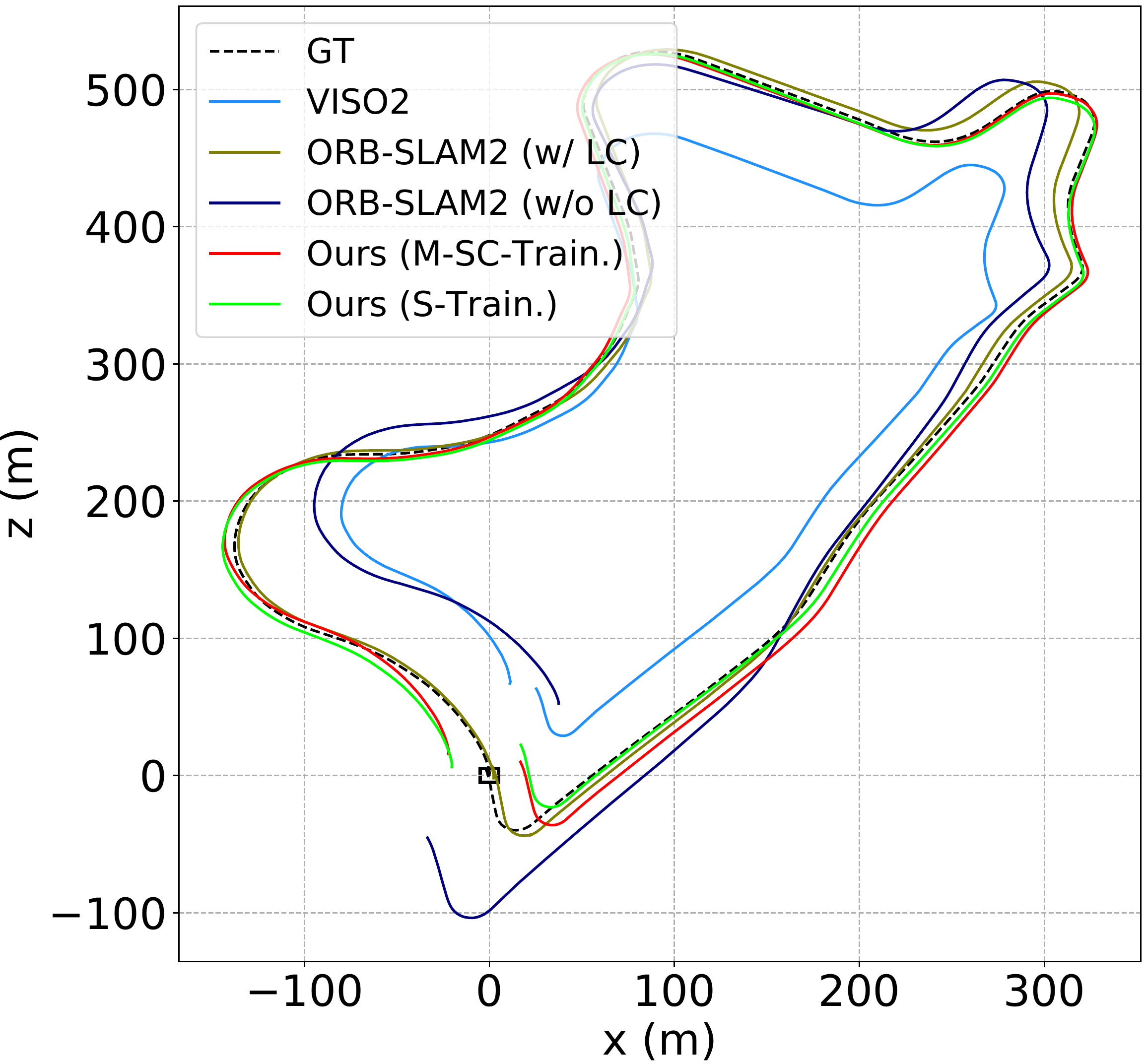}\\
        \end{multicols}
        \vspace{-20pt}
        \begin{multicols}{2}
            \centering
            \includegraphics[width=1\columnwidth]{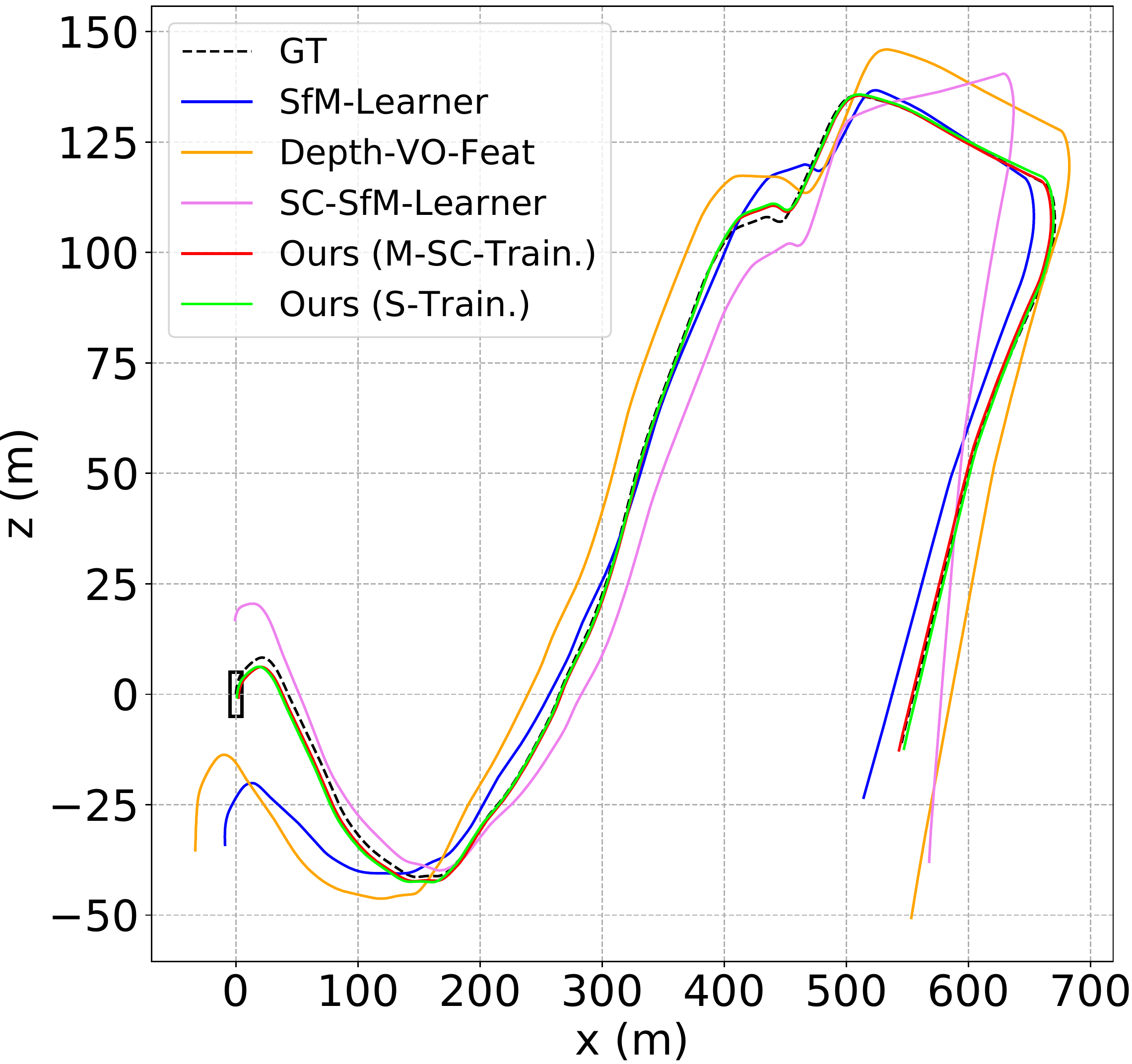}\\
            \includegraphics[width=1\columnwidth]{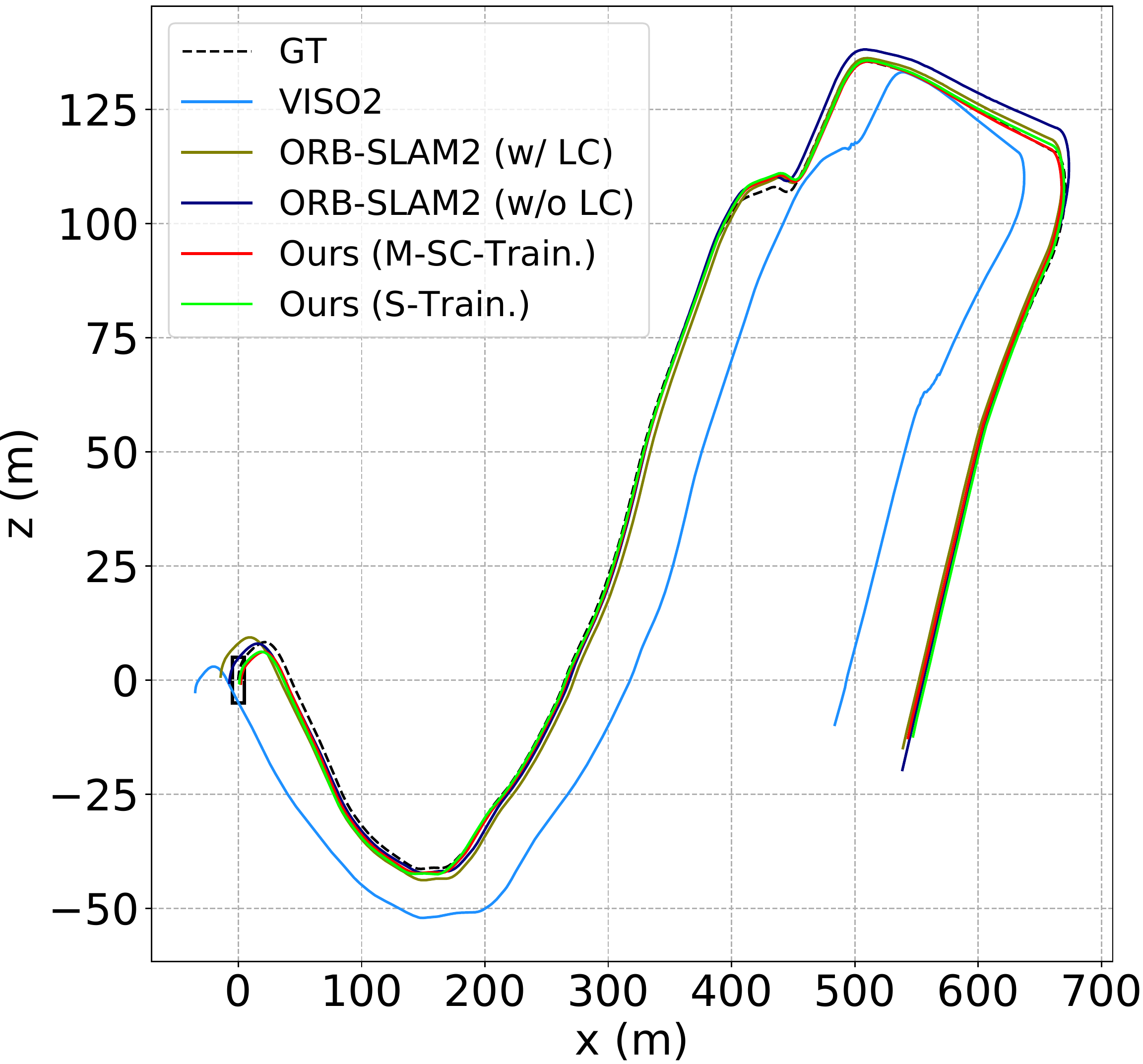}\\
        \end{multicols}
        \vspace{-20pt}
        \begin{multicols}{2}
            \centering
            \includegraphics[width=1\columnwidth]{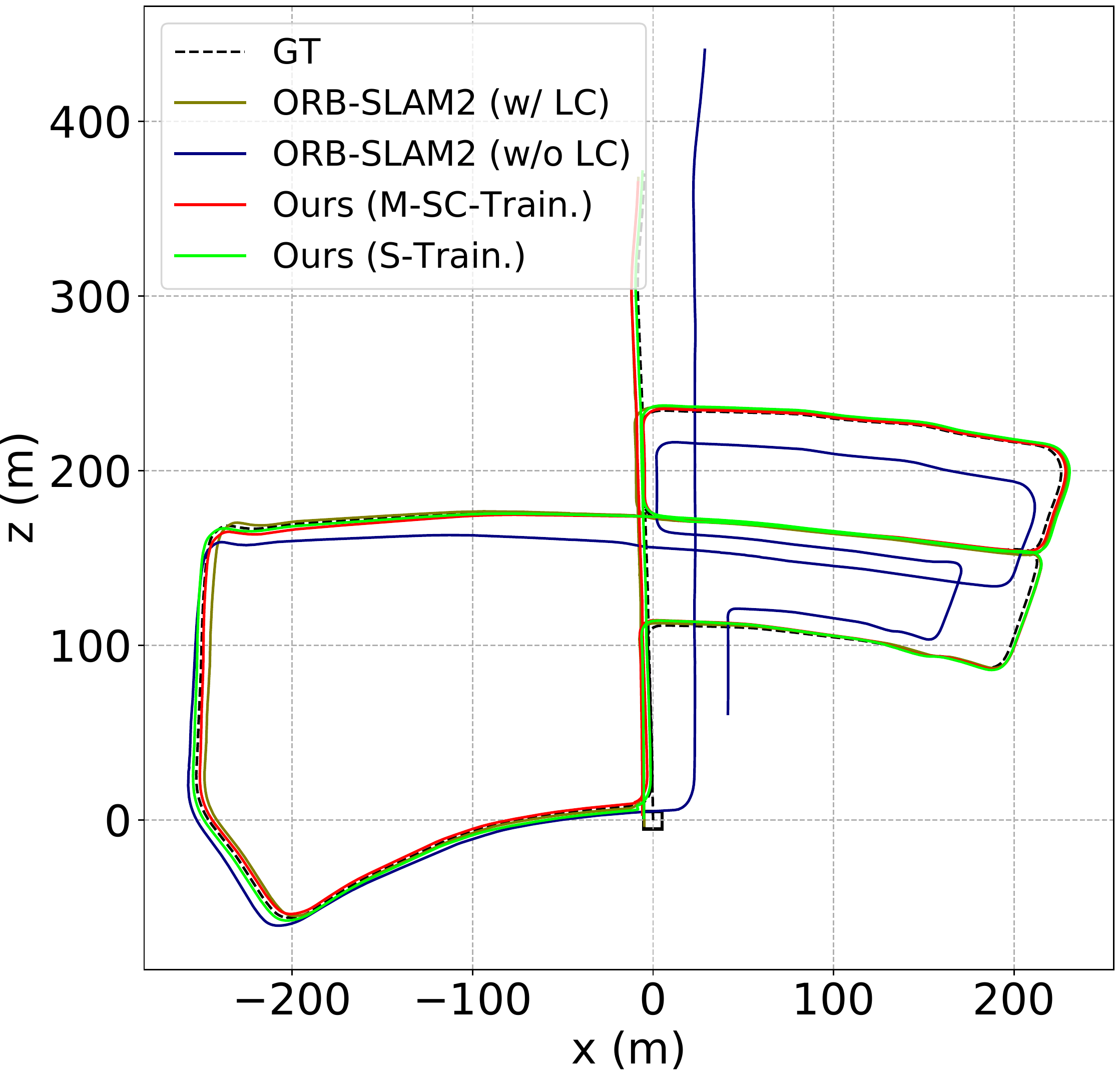}\\
            \includegraphics[width=1\columnwidth]{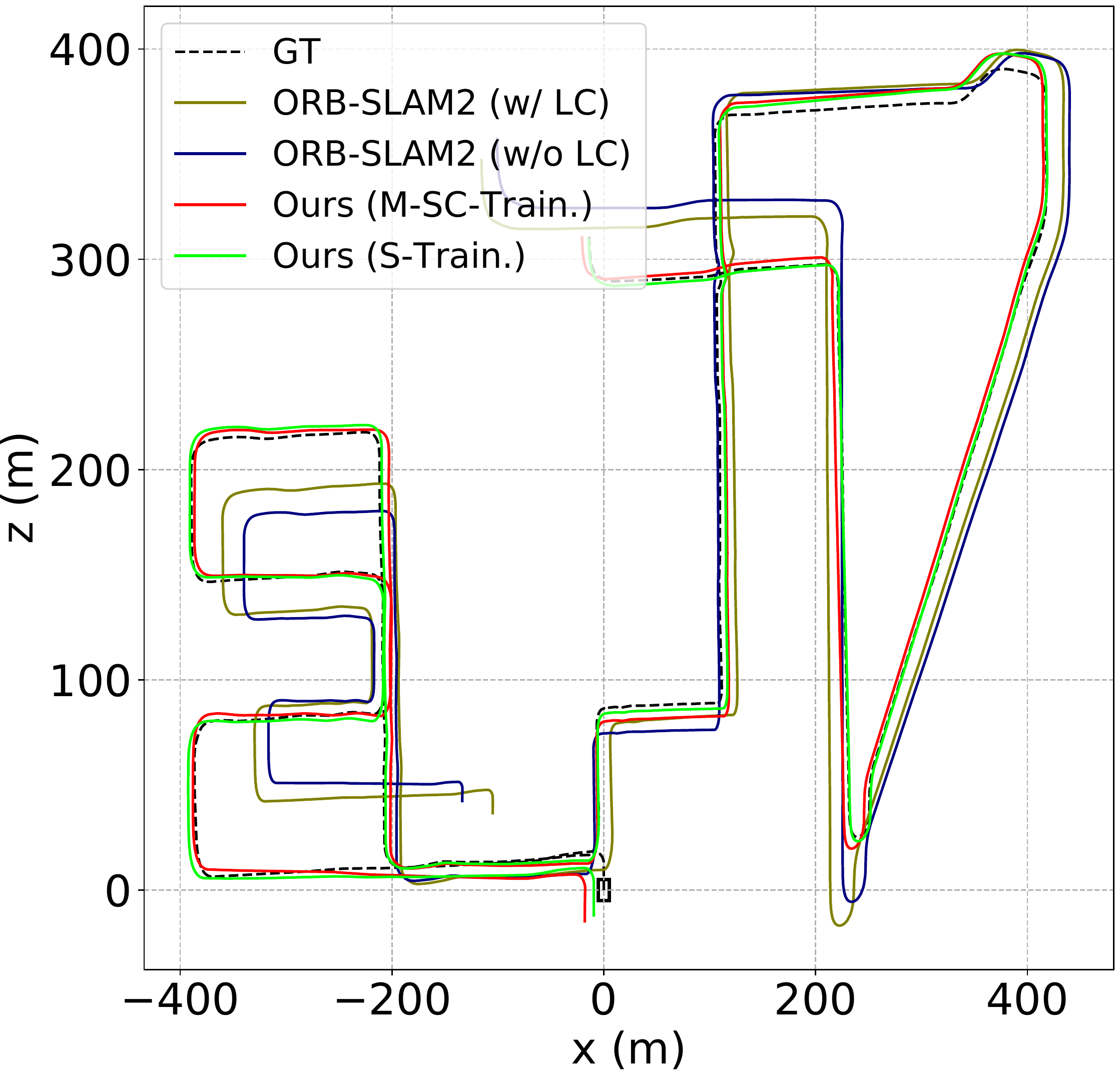}\\
        \end{multicols}
        \vspace{-20pt}
    \end{multirow}
\caption{Qualitative VO results on KITTI: (Top) Seq.09 and (Middle) Seq.10 against deep learning-based and geometry-based methods (shown separately) and
(Bottom) Seq.05 and Seq.08 against ORB-SLAM2. 
}
\vspace{-10pt}
\label{fig:traj}
\end{figure}

Depending on the availability of data (monocular/stereo videos, ground truth depths), different depth models can be trained. 
Throughout the work, we use a standard fully convolutional encoder-decoder network with skip-connections \cite{ronneberger2015u} \cite{monodepth2} to predict depths.

The most trivial way of training the depth model is using supervised learning \cite{eigen2014depth, liu2015depth, liu2016depth, laina2016deeper, kendall2017uncertainties, nekrasov2019multitask, fu2018dorn} but ground truth depths are not always available for different application scenarios. 
Recent works have suggested that learning single-view depths (and camera motions) in a self-supervised manner is possible using monocular sequences \cite{zhou2017sfmlearner, bian2019depth} or stereo sequences\cite{garg2016depth} \cite{godard2017monodepth} \cite{zhan2018depthVO} \cite{monodepth2}. 
Instead of using ground truth supervisions, the main supervision signal in the self-supervised frameworks is photometric consistency across multiple-views. 

In this work, we follow \cite{monodepth2} for training depth models using monocular sequences and stereo sequences. 
For self-supervised training, we jointly train the depth network and a pose network by minimizing the mean of the following \textit{per-pixel} objective function over the whole image. The \textit{per-pixel} loss is
\begin{align}
    L = &\min_{j}L_{pe}(I_{i}, I^{i}_{j})  + 
        \lambda_{ds} L_{ds}(D_i, I_i)  + \nonumber \\ 
        &\min_{j} \lambda_{dc} L_{dc}(D_{i}, D^{i}_{j}), \label{eqn:total_Loss}
\end{align}
where $[\lambda_{ds}, \lambda_{dc}]$ are loss weightings.
$L_{pe}$ is the photometric error by computing the difference between the reference image $I_i$ and the synthesized view $I^{i}_{j}$ from the source image $I_j$, where $j \in [i-n, i+n, s]$. 
$[i-n, i+n]$ are neighbouring views of $I_i$ while $s$ is stereo pair if stereo sequences are used in training. 
As proposed in \cite{monodepth2}, instead of averaging the photometric error over all source images, \cite{monodepth2} applies minimum to overcome the issues related to out-of-view pixels and occlusions. 
\begin{align}
    L_{pe}(I_{i}, I^{i}_{j}) &= 
    \frac{\alpha}{2}\left(1-\text{SSIM}(I_i, I^i_j)\right) + 
    (1-\alpha)|I_i-I^i_j| \\ 
    I^i_j &= f_{w}(I_j, h_1(K, D_i, T^j_i)) \label{eqn:photo_warp}, 
\end{align}
%
where
$\alpha=0.85$, 
$f_w(.)$ is a warping function. 
$K$ is the camera intrinsics. $D_i$ and $T^i_j$ are predicted from depth network and pose network respectively. 
$h_1(.)$ establish 
the reprojection from view-\textit{i} to view-\textit{j}.

$L_{ds}$ is an edge-aware depth smoothness for regularization.
\begin{align}
    L_{ds}(D_i, I_i) = 
        |\partial_x D_{i}| e^{-|{\partial_x I_{i}}|}+
        |\partial_y D_{i}| e^{-|{\partial_y I_{i}}|}, \label{eqn:depth_sm_Loss}
\end{align}
Similar to traditional monocular 3D reconstruction, scale ambiguity and scale inconsistency issues exist when monocular videos are used for training. 
Since the monocular training usually uses image snippets for training, it does not guarantee a consistent scale is used across snippets which creates scale inconsistency issue.
One solution is to use stereo sequences during training \cite{li2017undeepvo} \cite{zhan2018depthVO} \cite{monodepth2}, the deep predictions are aligned with real-world scale and scale consistent because of the constraint introduced by the known stereo baseline. 
Even though stereo sequences are used during training, only monocular images are required during inference.
Another possible solution to overcome the scale inconsistency issue is using temporal geometry consistency regularization proposed in \cite{zhan2019depthnormal} 
\cite{bian2019depth}, which constrains the depth consistency across multiple views. 
As depth predictions are consistent across different views, the scale inconsistency issue is resolved.
Using the rigid scene assumption as the cameras move in space over time we want the predicted depths at view-\textit{i} to be consistent with the respective predictions at view-\textit{j}.
This is done by \textbf{correctly transforming} the scene geometry from frame-$j$ to frame-$i$ much like the image warping.
Specifically, we adopt the inverse depth consistency proposed in \cite{zhan2019depthnormal}. 
%
\begin{align}
    L_{dc}(D_i, D^i_j) = |1/D_{i} - 1/D^i_j|
\end{align}
Inspired by \cite{monodepth2}, instead of averaging the depth consistency error over all source views, we use minimum to avoid occlusions and out-of-view scenes.

%
\begin{table*} [!t] 
\caption{
Quantitative result on KITTI Odometry Seq. 00-10. The best result is in bold and second best is underlined.
}
\begin{center}
\resizebox{2\columnwidth}{!}{%
\begin{tabular}{| c  | c | c c c c c c c c c c c | c |}
\hline
Method & Metric & 00 & 01 & 02 & 03 & 04 & 05 & 06 & 07 & 08 & 09 & 10 & Avg. Err.\\
\hline\hline
\multirow{5}*{SfM-Learner (updated ver.) \cite{zhou2017sfmlearner}} & 
$t_{err}$   & 
21.32 & \textbf{22.41} & 24.10 & 12.56 & 4.32 & 
12.99 & 15.55 & 12.61 & 10.66 & 11.32 & 15.25 & 14.068 \\ 
& $r_{err}$ & 
6.19 & 2.79 & 4.18 & 4.52 & 3.28 & 
4.66 & 5.58 & 6.31 & 3.75 & 4.07 & 4.06 & 4.660 \\ 
& ATE & 
104.87 & \underline{109.61} & 185.43 & 8.42 & 3.10 & 
60.89 & 52.19 & 20.12 & 30.97 & 26.93 & 24.09 & 51.701 \\ 
& RPE (m) & 
0.282 & \underline{0.660} & 0.365 & 0.077 & 0.125 & 
0.158 & 0.151 & 0.081 & 0.122 & 0.103 & 0.118 & 0.158 \\ 
& RPE ($^\circ$) &  
0.227 & 0.133 & 0.172 & 0.158 & 0.108 & 
0.153 & 0.119 & 0.181 & 0.152 & 0.159 & 0.171 & 0.160 \\ 
\hline 

\multirow{5}*{Depth-VO-Feat \cite{zhan2018depthVO}} & 
$t_{err}$   & 
6.23 & \underline{23.78} & 6.59 & 15.76 & 3.14 & 
4.94 & 5.80 & 6.49 & 5.45 & 11.89 & 12.82 & 7.911 \\ 
& $r_{err}$ & 
2.44 & 1.75 & 2.26 & 10.62 & 2.02 & 
2.34 & 2.06 & 3.56 & 2.39 & 3.60 & 3.41 & 3.470 \\ 
& ATE & 
64.45 & 203.44 & 85.13 & 21.34 & 3.12 & 
22.15 & 14.31 & 15.35 & 29.53 & 52.12 & 24.70 & 33.220 \\ 
& RPE (m) & 
0.084 & \textbf{0.547} & 0.087 & 0.168 & 0.095 & 
0.077 & 0.079 & 0.081 & 0.084 & 0.164 & 0.159 & 0.108 \\ 
& RPE ($^\circ$) &  
0.202 & 0.133 & 0.177 & 0.308 & 0.120 & 
0.156 & 0.131 & 0.176 & 0.180 & 0.233 & 0.246 & 0.193 \\ 
\hline 

\multirow{5}*{SC-SfMLearner \cite{bian2019depth}} & 
$t_{err}$   & 
11.01 & 27.09 & 6.74 & 9.22 & 4.22 & 
6.70 & 5.36 & 8.29 & 8.11 & 7.64 & 10.74 & 7.803 \\ 
& $r_{err}$ & 
3.39 & 1.31 & 1.96 & 4.93 & 2.01 & 
2.38 & 1.65 & 4.53 & 2.61 & 2.19 & 4.58 & 3.023 \\ 
& ATE & 
93.04 & \textbf{85.90} & 70.37 & 10.21 & 2.97 & 
40.56 & 12.56 & 21.01 & 56.15 & 15.02 & 20.19 & 34.208 \\ 
& RPE (m) & 
0.139 & 0.888 & 0.092 & 0.059 & 0.073 & 
0.070 & 0.069 & 0.075 & 0.085 & 0.095 & 0.105 & 0.086 \\ 
& RPE ($^\circ$) &  
0.129 & \textbf{0.075} & 0.087 & 0.068 & \underline{0.055} & 
0.069 & 0.066 & 0.074 & 0.074 & 0.102 & 0.107 & 0.083 \\ 
\hline 
\hline 

\multirow{5}*{VISO2 \cite{Geiger2011viso2}} & 
$t_{err}$   & 
10.53 & 61.36 & 18.71 & 30.21 & 34.05 & 
13.16 & 17.69 & 10.80 & 13.85 & 18.06 & 26.10 & 19.316 \\ 
& $r_{err}$ & 
2.73 & 7.68 & 1.19 & 2.21 & 1.78 & 
3.65 & 1.93 & 4.67 & 2.52 & 1.25 & 3.26 & 2.519 \\ 
& ATE & 
79.24 & 494.60 & 70.13 & 52.36 & 38.33 & 
66.75 & 40.72 & 18.32 & 61.49 & 52.62 & 57.25 & 53.721 \\ 
& RPE (m) & 
0.221 & 1.413 & 0.318 & 0.226 & 0.496 & 
0.213 & 0.343 & 0.191 & 0.234 & 0.284 & 0.442 & 0.297 \\ 
& RPE ($^\circ$) &  
0.141 & 0.432 & 0.108 & 0.157 & 0.103 & 
0.131 & 0.118 & 0.176 & 0.128 & 0.125 & 0.154 & 0.134 \\ 
\hline 

\multirow{1}*{DSO \cite{DSO} (from \cite{loo2018cnnsvo})} & 
ATE & 
113.18 & / & 116.81 & 1.39 & \textbf{0.42} & 
47.46 & 55.62 & 16.72 & 111.08 & 52.23 & 11.09 & 52.600 \\ 
\hline 

\multirow{5}*{ORB-SLAM2 (w/o LC) \cite{ORBSLAM2}} & 
$t_{err}$   & 
11.43 & 107.57 & 10.34 & \underline{0.97} & \underline{1.30} & 
9.04 & 14.56 & 9.77 & 11.46 & 9.30 & 2.57 & 8.074 \\ 
& $r_{err}$ & 
\underline{0.58} & \underline{0.89} & \textbf{0.26} & \textbf{0.19} & \underline{0.27} & 
\underline{0.26} & \underline{0.26} & 0.36 & \textbf{0.28} & \underline{0.26} & 0.32 & \underline{0.304} \\ 
& ATE & 
40.65 & 502.20 & 47.82 & \textbf{0.94} & 1.30 & 
29.95 & 40.82 & 16.04 & 43.09 & 38.77 & 5.42 & 26.480 \\ 
& RPE (m) & 
0.169 & 2.970 & 0.172 & 0.031 & 0.078 & 
0.140 & 0.237 & 0.105 & 0.192 & 0.128 & \textbf{0.045} & 0.130 \\ 
& RPE ($^\circ$) &  
0.079 & 0.098 & \underline{0.072} & 0.055 & 0.079 & 
\underline{0.058} & 0.055 & \underline{0.047} & 0.061 & \underline{0.061} & 0.065 & \underline{0.063} \\ 
\hline 

\multirow{5}*{ORB-SLAM2 (w/ LC) \cite{ORBSLAM2}} & 
$t_{err}$   & 
2.35 & 109.10 & \underline{3.32} & \textbf{0.91} & 1.56 & 
1.84 & \underline{4.99} & 1.91 & 9.41 & 2.88 & 3.30 & 3.247 \\ 
& $r_{err}$ & 
\textbf{0.35} & \textbf{0.45} & \underline{0.31} & \textbf{0.19} & \underline{0.27} & 
\textbf{0.20} & \textbf{0.23} & \underline{0.28} & \underline{0.30} & \textbf{0.25} & \textbf{0.30} & \textbf{0.268} \\ 
& ATE & 
\textbf{6.03} & 508.34 & \textbf{14.76} & \underline{1.02} & 1.57 & 
4.04 & 11.16 & 2.19 & 38.85 & \textbf{8.39} & 6.63 & 9.464 \\ 
& RPE (m) & 
0.206 & 3.042 & 0.221 & 0.038 & 0.081 & 
0.294 & 0.734 & 0.510 & 0.162 & 0.343 & \underline{0.047} & 0.264 \\ 
& RPE ($^\circ$) &  
0.090 & \underline{0.087} & 0.079 & 0.055 & 0.076 & 
0.059 & \underline{0.053} & 0.050 & 0.065 & 0.063 & 0.066 & 0.066 \\ 
\hline 

\multirow{1}*{CNN-SVO \cite{loo2018cnnsvo} (from \cite{loo2018cnnsvo})} & 
ATE & 
17.53 & / & 50.52 & 3.46 & 2.44 & 
8.15 & 11.51 & 6.51 & 10.98 & \underline{10.69} & 4.84 & 12.663 \\ 
\hline 

\multirow{5}*{\shortstack[1]{\textbf{Ours} \\\textbf{(Mono-SC Train.)}}} & 
$t_{err}$   & 
\underline{2.25} & 66.98 & 3.60 & 2.67 & 1.43 & 
\underline{1.15} & \textbf{1.03} & \textbf{0.93} & \underline{2.23} & \textbf{2.47} & \textbf{1.96} & \underline{1.972} \\ 
& $r_{err}$ & 
\underline{0.58} & 17.04 & 0.52 & 0.50 & 0.29 & 
0.30 & \underline{0.26} & 0.29 & \underline{0.30} & 0.30 & \underline{0.31} & 0.365 \\ 
& ATE & 
12.64 & 695.75 & 23.11 & 1.23 & 1.36 & 
\underline{3.75} & \underline{2.63} & \underline{1.74} & \underline{7.87} & 11.02 & \textbf{3.37} & \underline{6.872} \\ 
& RPE (m) & 
\underline{0.040} & 1.281 & \underline{0.063} & \underline{0.030} & \underline{0.057} & 
\underline{0.025} & \underline{0.033} & \underline{0.023} & \underline{0.042} & \textbf{0.055} & \underline{0.047} & \underline{0.041} \\ 
& RPE ($^\circ$) &  
\underline{0.056} & 0.725 & \textbf{0.046} & \underline{0.038} & \textbf{0.030} & 
\textbf{0.035} & \textbf{0.029} & \textbf{0.030} & \textbf{0.036} & \textbf{0.037} & \textbf{0.042} & \textbf{0.038} \\ 
\hline 

\multirow{5}*{\shortstack[1]{\textbf{Ours} \\\textbf{(Stereo Train.)}}} & 
$t_{err}$   & 
\textbf{1.96} & 56.76 & \textbf{2.38} & 2.49 & \textbf{1.03} & 
\textbf{1.10} & \textbf{1.03} & \underline{0.97} & \textbf{1.60} & \underline{2.61} & \underline{2.29} & \textbf{1.746} \\ 
& $r_{err}$ & 
0.60 & 13.93 & 0.55 & \underline{0.39} & \textbf{0.25} & 
0.30 & 0.30 & \textbf{0.27} & 0.32 & 0.29 & 0.37 & 0.364 \\ 
& ATE & 
\underline{11.34} & 484.86 & \underline{21.16} & 2.04 & \underline{0.86} & 
\textbf{3.63} & \textbf{2.53} & \textbf{1.72} & \textbf{5.66} & 10.88 & \underline{3.72} & \textbf{6.354} \\ 
& RPE (m) & 
\textbf{0.027} & 1.203 & \textbf{0.033} & \textbf{0.023} & \textbf{0.036} & 
\textbf{0.020} & \textbf{0.024} & \textbf{0.018} & \textbf{0.032} & \underline{0.056} & \underline{0.047} & \textbf{0.032} \\ 
& RPE ($^\circ$) &  
\textbf{0.055} & 0.773 & \textbf{0.046} & \textbf{0.037} & \textbf{0.030} & 
\textbf{0.035} & \textbf{0.029} & \textbf{0.030} & \underline{0.037} & \textbf{0.037} & \underline{0.043} & \textbf{0.038} \\ 
\hline

\end{tabular}
}
\end{center}
\vspace{-10pt}
\label{table:kitti_benchmark}
\end{table*}

\subsection{Optical Flow Network}
Many state-of-the-art deep learning based methods for estimating optical flow have been proposed \cite{dosovitskiy2015flownet, ilg2017flownet2,hui18liteflownet, sun2018pwc, meister2018unflow}.
In this work, we choose LiteFlowNet \cite{hui18liteflownet} as our backbone for $M_f$ as it is fast, lightweight, accurate and it has good generalization ability.
LiteFlowNet consists of a two-stream network for feature extraction and a cascaded network for flow inference and regularization. 
We refer readers to \cite{hui18liteflownet} for more details.
Even though LiteFlowNet \cite{hui18liteflownet} is trained with synthetic data (Scene Flow) \cite{dosovitskiy2015flownet}, it shows good generalization ability on real data.
In this work, we mainly use the LiteFlowNet model trained in Scene Flow.
However, we also show that a self-supervised training/finetuning can be performed which helps the model better adapt to unseen environments.
Similar to the self-supervised training of the depth network, the optical flow network is trained by minimizing the mean of the following \textit{per-pixel} loss function over the whole image,
\begin{align}\label{eqn:flow_Loss}
    L = &\min_{j}L_{pe}(I_{i}, I^{i}_{j})  + 
        \lambda_{fs} L_{fs}(||F^i_j||_2, I_i) \nonumber \\ 
        & + \lambda_{fc} L_{fc}(|-F^i_j - F^j_i |)  \\
    I^i_j &= f_{w}(I_j, h_2(F^i_j)), 
\end{align}
Different from Eqn. \ref{eqn:photo_warp}, $h_2(.)$ takes optical flow and establish the correspondences between view-\textit{i} and view-\textit{j}.
We also regularize the optical flow to be smooth using an edge-aware flow smoothness loss $L_{fs}(.)$.
Similar to Meister \etal \cite{meister2018unflow}, we estimate both forward and backward optical flow and constrain the bidirectional predictions to be consistent with the loss $L_{fc}$.

\section{Experiments and Evaluations} \label{sec:exp_eval}

In this section we describe the details of the experimental evaluation of our method. 
Due to space limitation, we mainly analyze our VO system on KITTI dataset \cite{Geiger2012kitti, Geiger2013kitti} and present indoor results in supplementary video.
We compare our system with prior arts on VO estimation. 
Additionally, we perform a detailed analysis to show the effect of different factors affecting the VO performance.

KITTI Odometry dataset contains 11 driving sequences (stereo sequences) with publicly available ground truth camera poses.
Following \cite{zhou2017sfmlearner}, we train our depth network and finetune our flow network on sequences 00-08. 
The dataset contains 36,671 training pairs, [$I_i, I_{i-1}, I_{i+1}, I_{i,s}$].

\subsection{Implementation Details}
We train our network with the PyTorch \cite{paszke2017pytorch} framework. 
All self-supervised experiments are trained using Adam optimizer \cite{kingma2014adam} for 20 epochs. 
For KITTI, images with size 640x192 are used for training.
Learning rate is set to $10^{-4}$ for the first 15 epochs and then is dropped to $10^{-5}$ for the remainder.
The loss weightings are $[\lambda_{ds}, \lambda_{dc}]=[10^{-3}, 5]$ and $[\lambda_{fs}, \lambda_{fc}]=[10^{-1}, 5 \times 10^{-3}]$ for single-view depth and optical flow experiments respectively.
\subsection{Visual Odometry Evaluation}
We compare our VO system with pure deep learning methods \cite{zhou2017sfmlearner }
\cite{zhan2018depthVO} \cite{bian2019depth}, geometry-based methods including DSO\cite{DSO}, VISO2\cite{Geiger2011viso2}, ORB-SLAM2\cite{mur2015orbslam} (w/ and w/o loop-closure) and CNN-SVO\cite{loo2018cnnsvo}.
The quantitative and qualitative results are shown in Table \ref{table:kitti_benchmark} and Fig. \ref{fig:traj}.
%
ORB-SLAM2 occasionally suffers from tracking failure or unsuccessful initialization. We run ORB-SLAM2 three times and choose the run with the least trajectory error.
\begin{table*}[th]  
\caption{
Ablation study: Effect on VO performance by changing different components of the Reference Model.
}
\begin{center}
\resizebox{0.8\textwidth}{!}{%
\begin{tabular}{| l  | c c c c | c c c c |}
\hline
\multirow{2}*{Experiment} &  
\multicolumn{4}{c|}{09} & 
\multicolumn{4}{c|}{10} \\
 {}  &
 $t_{err}$ & $r_{err}$  & RPE(m) & RPE ($^\circ$) &
 $t_{err}$ & $r_{err}$  & RPE(m) & RPE ($^\circ$) \\
\hline\hline

PnP only & 
7.12 & 2.43 & 0.082 & 0.081 &
6.83 & 3.88 & 0.058 & 0.093 \\
\hline

Depth model (Mono.)  & 
4.84 & 0.66 & 0.105 & 0.058 &
4.02 & 1.12 & 0.071 & 0.061 \\
\hline

Depth model (Mono-SC) & 
4.09 & 0.68 & 0.087 & 0.057 &
3.02 & 0.97 & 0.063 & 0.062 \\
\hline

Self-LiteFlowNet & 
2.97 & 0.64 & 0.048 & 0.047 &
2.87 & 0.77 & 0.047 & 0.053 \\
\hline

Uniform sampling &  
5.86 & 1.26 & 0.084 & 0.055 &
5.81 & 1.99 & 0.054 & 0.061 \\
\hline

\textit{Reference Model} (Sec.\ref{sec:ablation})  & 
4.56 & 0.62 & 0.082 & 0.056 &
3.53 & 1.14 & 0.054 & 0.060 \\
\hline
\hline

Full Image Res. & 
\textbf{2.61} & \textbf{0.29} & \textbf{0.056} & \textbf{0.037} &
\textbf{2.29} & \textbf{0.37} & \textbf{0.047} & \textbf{0.043} \\
\hline
\end{tabular}
}
\end{center}
\label{table:ablation}
\vspace{-20pt}
\end{table*}
We adopt common evaluation criteria for a more detailed analysis.
KITTI Odometry criterion evaluates possible sub-sequences of length (100, 200, ..., 800) meters and report the average translational error $t_{err} (\%)$ and rotational errors $r_{err} (^\circ/100m)$. 
Absolute trajectory error (ATE) measures the root-mean-square error between predicted camera poses $[x, y, z]$ and ground truth.
Relative pose error (RPE) measures frame-to-frame relative pose error. 
%
Since most of the methods are monocular method, which lacks a scaling factor to match with real-world scale, we scale and align (7DoF optimization) the predictions to the ground truth associated poses during evaluation by minimizing ATE \cite{umeyama1991least}.
Except for methods using stereo depth models (\textit{Ours (Stereo Train.)}, \textit{Depth-VO-Feat}) and known scale prior (VISO2), which already align predictions to real-world scale, for fair comparison, we perform 6DoF optimization w.r.t ATE instead.
We exclude Seq.01 while computing average error since a sub-sequence of Seq.01 does not contain trackable close features and most methods fail in this sequence. 
In Table \ref{table:kitti_benchmark}, \textit{Ours (Mono-SC Train.)} uses a depth model trained with monocular videos and inverse depth consistency for ensuring scale-consistency.
\textit{Ours (Stereo Train.)} uses a depth model trained with stereo sequences. Note that even stereo sequences are used in training, monocular sequences are used in testing. Therefore, \textit{Ours (Stereo Train.)} is still a monocular VO system.
We show that our method outperforms pure deep learning methods by a large margin in all metrics.
However, one interesting result for deep learning methods is that, 
all methods failed in Seq.01 except the deep models, which shows that deep models can be used as a complement to geometry methods with careful design in the future to overcome the failure mode for geometry methods.
For long sequence evaluation, ORB-SLAM2 shows less rotation drift $r_{err}$ but higher translation drift $t_{err}$ due to scale drift issue which is showed in Fig. \ref{fig:traj}. Sometimes the issue can be resolved by loop closing.
We use scale consistent depth predictions for scale recovery which mitigates the issue in most monocular VO/SLAM systems.
As a result, our method shows less translation drift over long sequences.
More importantly, our method shows consistently smaller RPE which allows our system to be a robust module for frame-to-frame tracking.

\subsection{Ablation Study} \label{sec:ablation}

Table \ref{table:ablation} shows an ablation study on our VO system.
We evaluate the effect of different components in our system by changing a variable in the \textit{Reference Model} consisting of:
\begin{itemize}
    \item Algorithm: Full algorithm shown in Alg. 1
    \item Depth model: Trained with stereo sequences
    \item Flow model: LiteFlowNet trained from synthetic dataset
    \item Matches: Top-\textit{N} matches with least flow inconsistency
    \item Image resolution: downsampled size ($640\times 192$)
\end{itemize}

\paragraph{Algorithm}
We compare the full algorithm with the algorithm using PnP only. 
Since the single-view depth predictions are not accurate enough to establish accurate 3D-2D matches, only using PnP performs worse than the full algorithm, which relies more on 2D-2D matches.

\paragraph{Depth model} 
We train three depth models with monocular or stereo sequences. 
For monocular experiments, depth models with and without depth consistency term, $\lambda_{dc}=[5, 0]$, are trained and used for depth estimation.
The result shows that with scale-consistent depth predictions, \textit{Mono-SC} model performs on par with the model trained with stereo sequences. 
However, both monocular models still have the scale ambiguity issue while \textit{Stereo Train.} does not.

\paragraph{Flow model}
Even LiteFlowNet is trained with synthetic data but it still shows good generalization ability from synthetic to real.
With self-supervised finetuning, the model adapts to driving sequences better and gives better VO result. 

\paragraph{Matches}
Forward-backward flow consistency as a measure to choose good 2D-2D matches is proposed.
To prove the effectiveness of the the flow consistency, we uniformly sample matches from the optical flows and compare the result against that using flow consistency.

\paragraph{Image resolution}
At inference time, simply increasing the image size to full resolution, the optical flow network predicts more accurate correspondences which helps the relative pose estimation.

\section{Conclusion} \label{sec:conclusion}
In this paper, we have presented a robust monocular VO system leveraging deep learning and geometry methods.
We explore the integration of deep predictions with classic geometry methods. 
Specifically, we use optical flow and single-view depth predictions from deep networks as an intermediate outputs to establish 2D-2D/3D-2D correspondences for camera pose estimation.
We show that the deep models can be trained in a self-supervised manner and we explore the effect of different training schemes.
Depth models with consistent scale can be used for scale recovery, which mitigates the scale drift issue in most monocular VO/SLAM systems.
Instead of learning a complete VO system in an end-to-end manner, which does not perform competitively to geometry-based methods, 
we think that (1) integrating deep predictions with geometry-based methods can gain the best from both domains; 
(2) deep VO models can be used as a complement in standard VO/SLAM system when they fail.
A future work will be turning the proposed system into a map-to-frame tracking system to further improve the tracking performance.


\section{Acknowledgement}
This work was supported by the UoA Scholarship to HZ, the ARC Laureate Fellowship FL130100102 to IR and the Australian Centre of Excellence for Robotic Vision CE140100016.

\clearpage

\end{document}